%% file: main.tex
\documentclass[runningheads]{llncs}
\input{packages}
\input{definitions}
\begin{document}
\title{A Claim Decomposition Benchmark for Long-form Answer Verification}
%
%
\author{Zhihao Zhang \and
Yixing Fan \and
Ruqing Zhang \and
Jiafeng Guo}
%
\authorrunning{Z. Zhang et al.}
%
\institute{Institute of Computing Technology, Chinese Academy of Sciences,  Beijing, China \\
\email{\{zhangzhihao22s,fanyixing,zhangruqing,guojiafeng\}@ict.ac.cn}}
%
\maketitle              
\begin{abstract}
The advancement of \acp{LLM} has significantly boosted the performance of complex long-form question answering tasks. However, one prominent issue of \acp{LLM} is the generated ``hallucination'' responses that are not factual. 
Consequently, attribution for each claim in responses becomes a common solution to improve the factuality and verifiability.
Existing researches mainly focus on how to provide accurate citations for the response, which largely overlook the importance of identifying the claims or statements for each response.
To bridge this gap, we introduce a new claim decomposition benchmark, which requires building system that can identify atomic and checkworthy claims for \ac{LLM} responses.
Specifically, we present the \ac{CACDD}, which builds on the WebCPM dataset with additional expert annotations to ensure high data quality. 
The \ac{CACDD} encompasses a collection of 500 human-annotated question-answer pairs, including a total of 4956 atomic claims. We further propose a new pipeline for human annotation and describe the challenges of this task. 
In addition, we provide experiment results on zero-shot, few-shot and fine-tuned \acp{LLM} as baselines. The results show that the claim decomposition is highly challenging and requires further explorations.
All code and data are publicly available~\footnote{https://github.com/FBzzh/CACDD}. 

\end{abstract}
\section{Introduction}

In recent years, \acp{LLM} have demonstrated excellent performance in various domains of \ac{NLP} due to their robust natural language capabilities. Through training on extensive data with a large number of parameters,  these models have shown significant advancements in their ability to understand and process natural language. One of the most notable improvements in \acp{LLM} is their ability for complex reasoning and the generation of lengthy, coherent responses, which has greatly enhanced their capability to answer complex, long-form questions. However, the powerful generation capabilities of these models have also led to the emergence of the so-called "hallucination" problem, where the models occasionally generate content that does not align with reality. This issue not only affects the reliability of the models but also significantly limits their use in real-world applications, especially in high-stakes, risk-sensitive tasks where the accuracy of factual information is crucial.


In response to the issue of hallucination generation in large language models, researchers have proposed a variety of solutions to verify the credibility of the generated response, including multi-path result cross-validation, context consistency verification, and external knowledge verification. Initially, multi-path result cross-validation identifies and reduces potential hallucinations by comparing the output of different models or different parameter settings of the same model. For instance, Wang et al.\cite{wang2023self} proposed the Self-Consistency method, which selects the most consistent answer by sampling different reasoning paths. Subsequently, context consistency verification focuses on ensuring that the text generated by the model is logically and semantically consistent with its input context, thereby reducing the generation of hallucinations. For example, Shi et al.\cite{shi2024trusting} introduced Context-Aware Decoding (CAD), which amplifies the probability difference of the model's output with and without context, making the model more context-compliant. Lastly, external knowledge verification enhances the factual accuracy of the generated responses by comparing them with external knowledge bases or retrieved evidence documents. For instance, Gao et al.\cite{gao2023rarr} proposed the Retrofit Attribution using Research and Revision (RARR), which reduces hallucinations in large model responses by detecting and modifying content inconsistent with retrieved relevant evidence. 


Despite the fact that current research has proposed a variety of answer verification approaches from different perspectives, these approaches mainly focus on verifying the consistency between the response and the facts, often neglecting the importance of identifying and decomposing the claims that are worthy to be verified in the results. For instance, Yue et al.\cite{yue2023automatic} regards the entire response as a single claim and compares it with the reference. While Gao et al.\cite{gao2023enabling} assume that each natural sentence within the response constitutes an independent claim, and use the NLTK tool to identify sentences, subsequently comparing each sentence to the reference crafted by human experts. These approaches have not delved into the intrinsic nature of the claims, thereby constraining the accuracy and reliability of the answer verification.


\begin{figure}[htbp]
\includegraphics[width=0.9\textwidth]{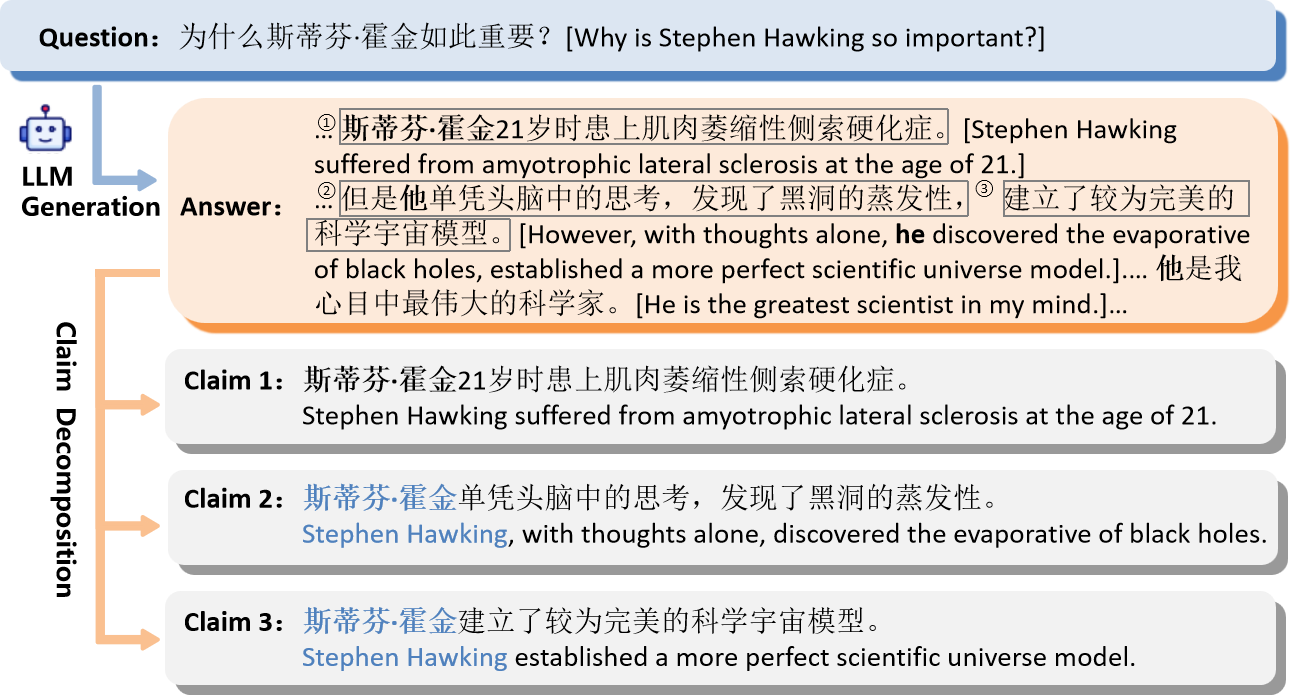}
\caption{A example of the claim decomposition.} \label{fig11}
\end{figure}

In reality, the factual verification of long response is an extremely challenging task. This is primarily due to the fact that a lengthy response typically encompasses one or more claims, with complex co-reference relationships existing among these claims. As shown in Figure\ref{fig11}, the answer to the question "Why is Stephen Hawking so important?" includes three claims, and there are referential relationships between claims 2 and 3 and claim 1. Additionally, the sentence "He is the greatest scientist in my mind" in the answer does not contain valuable information and does not require verification. It is evident that the verification of responses from \acp{LLM} usually requires: 1) identifying claims; 2) atomic decomposition of claims; 3) factual verification of each claim. Consequently, accurately identifying the claims in the response and decomposing each claim into atomic claims is an important component in the effective answer verification.


To address the tasks mentioned, this paper introduces the first benchmark for Chinese claim decomposition in answer verification, aiming to enhance the reliability of \acp{LLM} in answering complex questions. Initially, we define the concept of atomic claims, inspired by Russell's philosophy of logical atomism~\cite{russell1918philosophy}, which encompasses four fundamental principles: Indivisibility, Semantic Integrity, Verifiability, and Context Independence. Subsequently, we design an annotation pipeline for atomic claim decomposition. Based on this pipeline, we have created the \ac{CACDD} dataset through manual annotation, which is built on the WebCPM dataset. Each claim in this dataset has been resolved for co-references to satisfy the principle of context independence. In total, the CACDD includes 500 questions and 4956 claims, with all the annotated data and the code publicly accessible on \url{https://github.com/FBzzh/CACDD}. 


In order to better understand the claim decomposition task, we have conducted experimental analyses to assess the the performance of \acp{LLM} on this benchmark. Specifically, we have conducted zero-shot and few-shot experiments on open-source \acp{LLM}, as well as GPT-3.5. The results indicate that the open-source \acp{LLM} do not perform satisfactorily on this task, and even the outstanding GPT-3.5 still has a significant gap compared to humans.

\par The main contributions of this study include: 
\begin{enumerate}
    \item We have constructed the first Chinese Atomic Claim Decomposition Dataset (CACDD) for claim decomposition task, providing a valuable resource for future research; 
    \item We have tested the performance of the claim decomposition task on several advanced \acp{LLM}, providing a foundation for further optimization; 
    \item We have provided a clear definition of atomic claims, offering a standard that can be followed for the claim decomposition task. 
\end{enumerate}

\section{Related Work}
\subsection{Claim Decomposition Methods}
\subsubsection{Lexical Parsing Methods}
The traditional Claim decomposition task aims to identify the atomic claims within complex sentences, which is commonly used for tasks such as summarization, argument mining, and question answering. Before the advent of large language models, most related research was based on lexical parsing methods. The DCP~\cite{gao2019automated} method extracts the verb phrase components and clauses from complex sentences as criteria for summarization tasks. The DisSim~\cite{niklaus2019dissim} framework transforms complex sentences into a more regular intermediate representation by dividing them into a two-layered semantic hierarchy consisting of core facts and accompanying context. Moreover, the PredPatt~\cite{white2016universal} framework decomposes complex sentences by extracting predicate argument structures from syntactic dependency parsing based on the Universal Dependencies~\cite{de2021universal} Project. However, these methods based on lexical parsing typically only generate a structured representation of claims, rather than complete and coherent sub-claims, thereby constraining their application in downstream NLP tasks such as fact-checking and answer verification.
\subsubsection{LLM Prompting Approaches}
Upon their emergence, LLMs have garnered widespread attention across various domains of natural language processing due to their outstanding natural language comprehension and generation capacities. Recent research pertaining to claim decomposition tasks has also capitalized on the language ability of LLMs. Most of these LLM-based approaches are implemented through prompts~\cite{chen2024sub,jing2023faithscore,kamoi2023wice,min2023factscore,wang2023factcheck}. By providing examples that are carefully crafted by humans, they instruct LLMs to complete this task through in-context learning. In contrast to lexical parsing methods, these LLM prompting approaches are more flexible and unstructured, enabling the generation of coherent and fluent claims. Nonetheless, the LLM prompting approaches are also affected by the hallucinations inherent in LLMs, which can result in the generation of sub-claims that deviate from the original text. Furthermore, to ensure the quality of claim decomposition, researchers are required to provide enough in-context examples, leading to the prompts often being excessively lengthy. The excessive length of these prompts can lead to a reduction in inference efficiency and an increase in token costs.
\subsection{Answer Verification}
The verification of the extent to which LLM-generated responses are supported by the context or retrieved evidence is highly dependent on the claim decomposition task.  FactScore~\cite{min2023factscore} evaluates the factual precision of the LLM-generated responses through decompose them into sub-claims and verify each claim using a knowledge source. Factcheck-GPT~\cite{wang2023factcheck} introduces an end-to-end annotation framework that assesses the factual accuracy of LLM-generated responses through a combined approach that involves both LLM and human evaluation. Chen~\cite{chen2022generating} employs problem decomposition to decompose LLM-generated responses, verifying complex claims from various aspects, both explicit and implicit, by generating a series of yes or no sub-questions. However, the datasets for these works are primarily derived from fact-intensive domains such as biographies~\cite{min2023factscore} and political claims~\cite{chen2024complex,chen2022generating}, exhibiting a deficiency in data and pertinent research from long-form question-answering scenarios.
\section{The Atomic Claim Decomposition Task}
In this section, we first introduce the definition of atomic claim along with its respective characteristics, followed by an exposition of the definition and challenges associated with the claim decomposition task.
\subsection{Atomic Claim Definition}
The definition of atomic of Claims has long been a contentious issue. Wang~\cite{wang2023factcheck} believes that it is challenging to define atomic and then determine the granularity of decomposition. Unfortunately, their discourse on atomic did not arrival at a definitive conclusion.  Wanner~\cite{wanner2024closer} traces this issue back to the philosophical level. They employ Bertrand Russell's philosophy of logical atomism~\cite{russell1918philosophy} and the neo-Davidsonian analysis~\cite{lemmon1967comments} as the theoretical guidance, and manually decomposed 21 examples as the guidance for LLMs to perform the claim decomposition task. 
\par We also fellow the guidance of Bertrand Russell's philosophy of logical atomism and provide the following definition for an atomic claim:
\begin{definition}
An atomic claim delineates an indivisible minimal fact, which either describes a property of an individual or the relationship between individuals.
\end{definition}
An atomic claim should possess the following properties:
\begin{description}
    \item [Indivisibility] An atomic claim should delineates a single, indivisible fact, which means that it cannot be decomposed into simpler, more fundamental claims without loss of meaning.
    \item [Semantic integrity] An atomic claim should contain sufficient information to preclude any inconsistencies or ambiguities with the original claim.
    \item [Verifiability] An atomic claim should be verifiable. This means that it should delineates a fact that is worthy to be verified, rather than a commonsense or subjective opinion.
    \item [Context independence] An atomic claim should be context-independent, therefore its truthfulness can be assessed independently without preceding and following context.
\end{description}
\subsection{Atomic Claim Decomposition Task}
This study primarily focuses on the \ac{LFQA} task. Long-form question answering  aims at answering complex, open-ended questions with detailed, paragraph-length responses\cite{qin2023webcpm}. We define the task of atomic claim decomposition within this scenario. Given a question $q$ and a long response $r$ generated by \acp{LLM}, the goal is to decompose the response into a list of atomic claims $C = \{ac_1, ac_2, ..., ac_n\}$. Assuming the decomposition method is denoted by $D$, this can be formulated into the following function:
\begin{equation}
D(q,r) = \{ac_1, ac_2, ..., ac_n\}
\end{equation}
\par It should be noted that only the facts within the response require decomposition. Hence, the decomposition method must be capable of identifying non-factual sentences within the response and disregarding them.

\section{Chinese Atomic Claim Decomposition Dataset}
\begin{figure}[!t]
\includegraphics[width=0.9\textwidth]{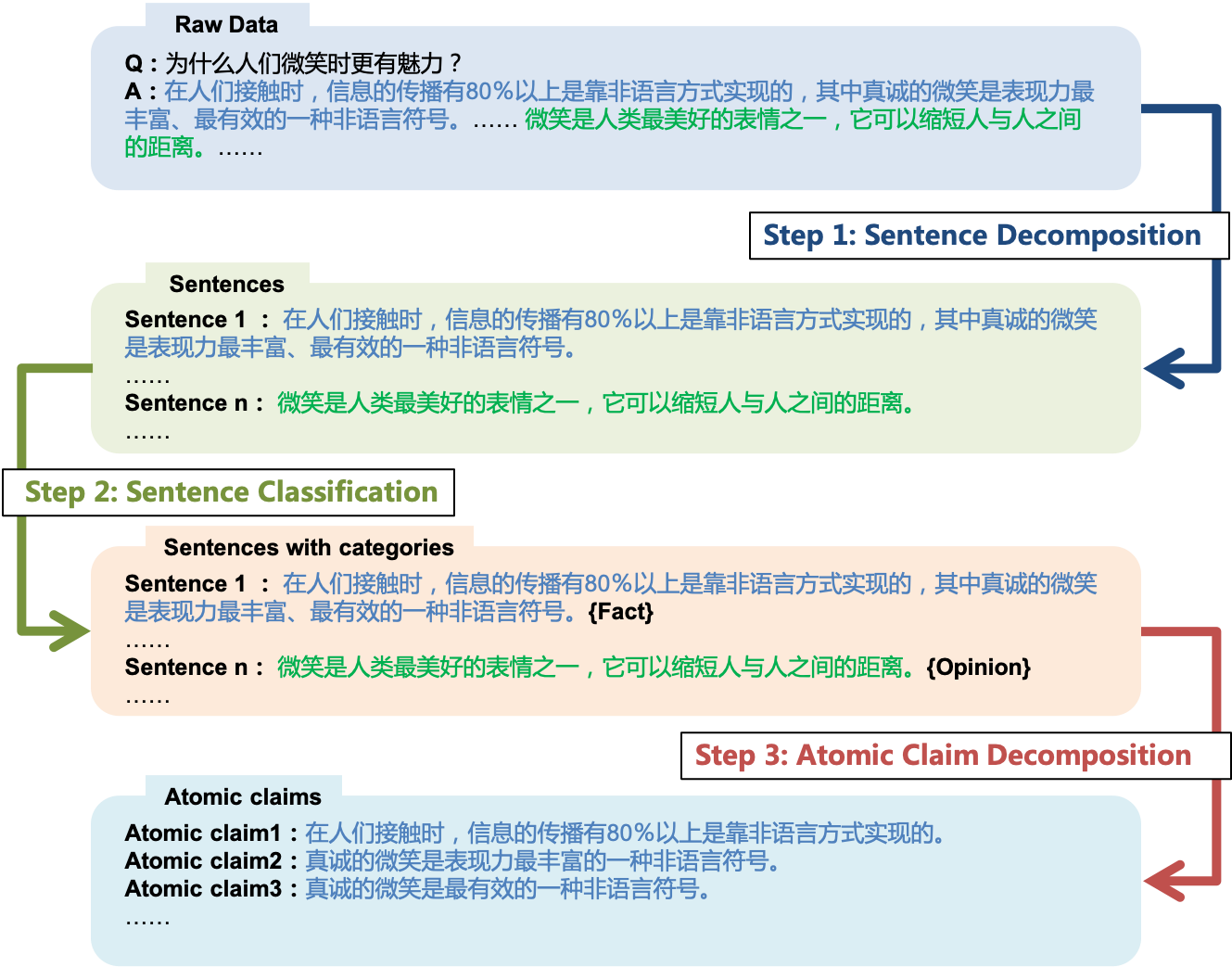}
\caption{A example of the data annotation pipeline.} \label{fig0}
\end{figure}

To facilitate research on atomic claim decomposition task, we introduce Chinese Atomic Claim Decomposition Dataset, a dataset designed for claim decomposition task within the \ac{LFQA} scenario. We first select data from WebCPM~\cite{qin2023webcpm} dataset and then annotate them by our human-annotation pipeline, which includes three steps: sentence decomposition, sentence classification and atomic claim decomposition. Figure~\ref{fig0} shows a example of our data annotation pipeline.

\subsection{Data Collection}
WebCPM, an innovative open-source project, is designed to advance the field of interactive search research. Utilizing Chinese pre-trained models, this project imitates human web search behaviors and answers questions based on the facts collected from the websites it returns. It introduces a dataset that contain questions and \ac{LLM} generated answers, which is suitable for our task. Therefore, We sample the first ten percent (550 of 5500) of WebCPM dataset to serve as the foundation for our subsequent manual annotation.
\subsection{Data Annotation}
Our data annotation pipeline can be divided into three parts. First, we decompose the whole response into sentences. Then, we classify each sentence into four categories. Finally, for sentences that have been labeled as fact, we decompose them into atomic claims. 

\subsubsection{Sentence Decomposition}
Given a response generated by \acp{LLM}, it is infeasible to directly decompose it into atomic claims. Thus, we first decompose it into nature language sentences using \ac{NLTK}, which allows annotators to focus more on the decomposition of atomic claims within individual sentences, thereby reducing the difficulty of annotation. Along with sentence decomposition, we also filter out the citation marks in the responses.

\subsubsection{Sentence Classification}
Not all claims in a response are worthy to be verified. For example, subjective opinions like \textit{"I think the iPad 10 has a very high cost performance ratio."} or instructions like \textit{"Don't blindly follow the trend anymore."} do not need to check their truthfulness. A claim is considered checkworthy if it is one for which the general public has an interest in knowing the truth~\cite{hassan2015detecting}. However, we assume that people who ask \acp{LLM} are interested in all factual claims in the response.
\par Considering the characteristics of \ac{LFQA} data, we classify the sentences into four categories: fact, opinion, instruction and others. For each category, we provide a definition along with the subcategories it encompasses.
\begin{description}
    \item [Fact] A fact is defined as a description of an objectively existing entity or event. Facts are objective, verifiable, and not influenced by personal beliefs or opinions. For example, "the main destructive power of nuclear bombs comes from the shock wave effect." is a fact.
    \item [Opinion] A opinion is defined as a belief or judgment that is not necessarily based on absolute certainty or proof. It is a personal perspective or view that may be influenced by subjective interpretations, emotions, or biases. For example, "Stephen Hawking is the greatest scientist in history." is a opinion.
    \item [Instruction] An instruction refers to a directive or command that specifies how a procedure or process should be executed. It outlines the actions to be taken to achieve a particular outcome or to operate a system or device. For example, "Don't blindly follow the trend anymore." is a instruction.
    \item [Other] This category contains sentences that cannot be classified into the three categories mentioned above, such as sentences that are rhetorical questions or connect the context. 
\end{description}

\subsubsection{Atomic Claim Decomposition}
As mentioned above, we only perform atomic claim decomposition on sentences that are classified as facts in the previous step. Following the definition of atomic claims outlined in Section 3.1, factual sentences are decomposed into context-independent, verifiable atomic claims. During the decomposition process, annotators can simultaneously view the sentence to be decomposed along with the preceding and following sentences, as well as the questions, which provide both local and global context for the decomposition. This is beneficial for annotators to perform anaphora resolution.

\subsection{Dataset Analysis}

\begin{figure}[htbp]
\vspace{-2.0em}
\resizebox{0.9\textwidth}{!}{
\includegraphics[width=1.0\textwidth]{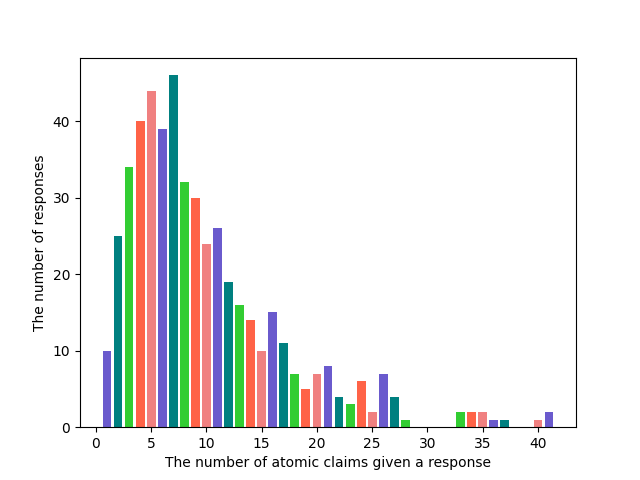}
}
\caption{The distribution of atomic claims amount given a response.} \label{fig1}
\vspace{-3.0em}
\end{figure}

\subsubsection{Statistics} 


We have retained 500 question-answer pair data containing a list of atomic declarations, totaling 4,956 atomic claims. The average atomic claims of each response is 9.912, which means that, on average, each response contains nearly ten distinct atomic claims. Figure~\ref{fig1} provides insight into the density of facts within the responses. It also suggests that the responses are rich in detail and that the decomposition process effectively captures the multiple atomic claims within each factual sentences.

\subsubsection{Sentences}
\begin{figure}[htbp]
  \centering
  \vspace{-2.0em}
  \begin{minipage}[b]{0.48\textwidth}
    \includegraphics[width=\textwidth]{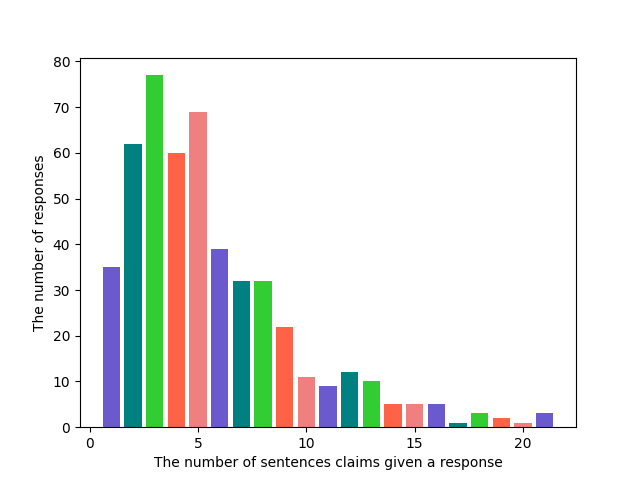}
    \caption{The distribution of sentences amount given a response.}
    \label{fig2}
  \end{minipage}
  \hfill 
  \begin{minipage}[b]{0.48\textwidth}
    \includegraphics[width=\textwidth]{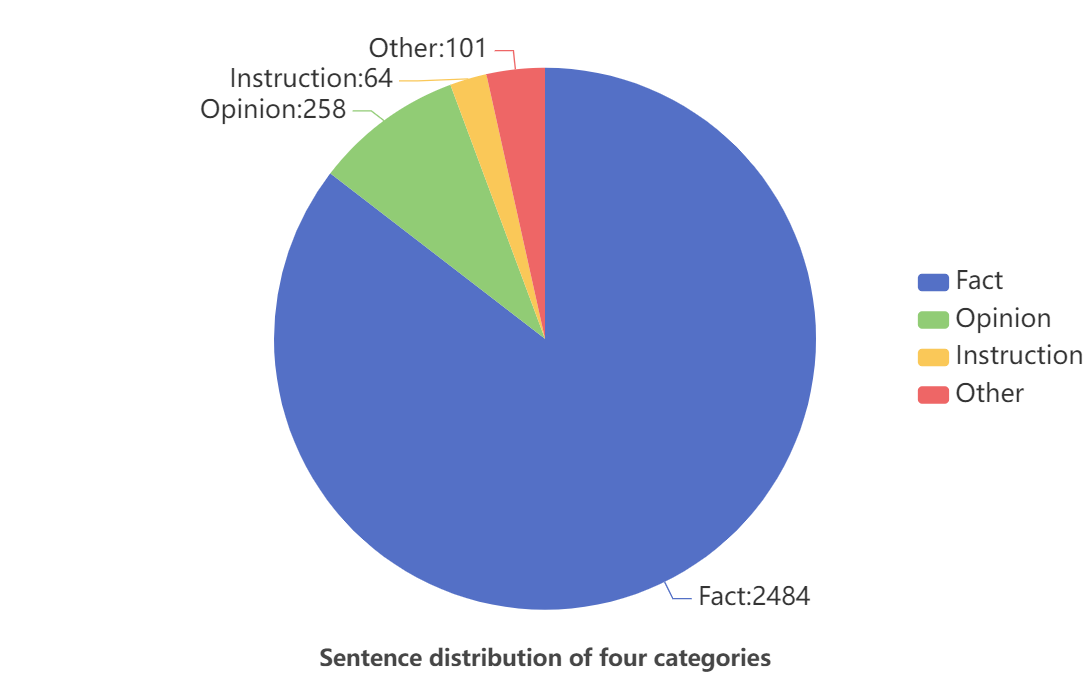}
    \caption{The distribution of sentences categories.}
    \label{fig3}
  \end{minipage}
  \vspace{-3.0em}
\end{figure}

Most responses contain less than 10 sentences, while the longest response contain 39 sentences and there are 5 responses contain more than 21 sentences. Figure~\ref{fig2} shows the number of responses which contain less than 21 sentences. In addition, as shown in Figure~\ref{fig3}, most of the sentences are facts, totaling 2,484 of 2,907 sentences. 258 sentences are considered to be opinion and 64 are instruction. 

\section{Experiments}
\subsection{Experimental Setup}
We use \ac{CACDD} dataset for atomic claim decomposition task with multi advanced \acp{LLM}. Open source \acp{LLM} include Baichuan2-7B, Glm4-9B, Llama3-8B, Mistral-7B, Qwen2-7B and Solar-10.7B. Moreover, we also measure the performance of GPT3.5-turbo. For better instruction compliance, we choose the chat or instruct version of the \acp{LLM} mentioned above. We prompt these \acp{LLM} under two settings: zero-shot and three-shot. For zero-shot, we just introduce the atomic claim definition to the model and instruct it to decompose the whole response in the prompt. For three-shot, we carefully selected three examples as context information for models to learn. In detail, we selected examples that require anaphora resolution or ellipsis supplementation from the context of the question or answer, and contain multiple factual and non factual sentences at the same time. All prompts can be found in our github repository. To facilitate reproducibility, all of our experiments set the "do\_sample" parameters to false and employ greedy decoding. 
We also fine-tune a Llama3-8B model using 350 data from the \ac{CACDD} dataset as strong baseline.
\subsection{Metrics}
For evaluation, we report the precision, recall and F1 scores of three widely-used metrics of text similarity: EM, Rouge-l and BertScore. For the Rouge-l metric, we set the threshold for text match to 0.8, while for the BertScore metric it is 0.9.

\subsection{Results}

\begin{table}[htbp]
\centering
\resizebox{\textwidth}{!}{
\setlength{\tabcolsep}{1mm}{
\begin{tabular}{*{10}{c}}
  \toprule
    \multirow{2}*{Model} & \multicolumn{3}{c}{EM} & \multicolumn{3}{c}{Rouge-l} & \multicolumn{3}{c}{BertScore} \\
  \cmidrule(lr){2-4}\cmidrule(lr){5-7}\cmidrule(lr){8-10}
    & Precision & Recall & F1 & Precision & Recall & F1 & Precision & Recall & F1  \\
  \midrule
        Baichaun2-7B & \textbf{9.28} &5.74 &6.25 &29.68 &18.35 &20.30 &29.87 &18.53 &20.56 \\
        Glm4-9B      & 4.89 &6.00 &5.21 &\textbf{31.57} &\textbf{39.70} &\textbf{33.30} &\textbf{30.68} &\textbf{38.55} &\textbf{32.38} \\
        Llama3-8B    & 0.17 &0.44 &0.23 &16.21 &27.41 &18.98 &15.06 &25.16 &17.56 \\
        Mistral-7B   & 2.64 &2.79 &2.53 &18.75 &18.78 &17.75 &20.47 &20.51 &19.37 \\
        Qwen2-7B     & 1.88 &2.19 &1.88 &27.90 &27.02 &26.04 &28.72 &27.65 &26.79 \\
        Solar-10.7B  & 0.02 &0.03 &0.02 &0.50 &0.77 &0.56 &0.61 &0.92 &0.68 \\
        GPT3.5-turbo & 7.98 &\textbf{9.70} &\textbf{8.26} &27.23 &31.75 &27.64 &28.36 &32.48 &28.45 \\
  \bottomrule
\end{tabular}}}
    \caption{Zero-shot results of \acp{LLM} on atomic claim decomposition.}
    \label{tab2}
\vspace{-3.0em}
\end{table}

Table~\ref{tab2} and \ref{tab3} show the results of our experiments. For zero-shot setting, Chinese open-source \acp{LLM}, including Glm4-9B, Baichuan2-7B and Qwen2-7B, show their better performance on this task. Glm4-9B achieves the best results on the Rouge-l and BertScore metrics, demonstrating its strong Chinese comprehension ability. The reason is that these models are enhanced for Chinese data, and can better understand and follow Chinese prompt. However, the performance of models that are not enhanced perform poorly on zero-shot setting. For example, the performance of Solar-10.7B is poor, as its responses contain a significant number of English sentences, as well as Chinese sentences that interspersed with Korean phrases. GPT3.5 achieves the best results on EM metric, which shows its excellent Chinese understanding and language ability.Nevertheless, the performance is not satisfactory yet.

\begin{table}[htbp]
\centering
\vspace{-2.0em}
\resizebox{\textwidth}{!}{
\setlength{\tabcolsep}{1mm}{
\begin{tabular}{*{10}{c}}
  \toprule
    \multirow{2}*{Model} & \multicolumn{3}{c}{EM} & \multicolumn{3}{c}{Rouge-l} & \multicolumn{3}{c}{BertScore} \\
  \cmidrule(lr){2-4}\cmidrule(lr){5-7}\cmidrule(lr){8-10}
    & Precision & Recall & F1 & Precision & Recall & F1 & Precision & Recall & F1  \\
  \midrule
        Baichaun2-7B & 15.72 &12.91 &13.26 &38.73 &31.04 &32.43 &39.20 &31.55 &32.89 \\
        Glm4-9B      & 9.32 &11.15 &9.71 &35.14 &41.87 &36.51 &36.68 &43.83 &38.12 \\
        Llama3-8B    & \textbf{20.28} &\textbf{23.04} &\textbf{20.62} &\textbf{43.67} &\textbf{48.73} &\textbf{44.09} &\textbf{43.89} &\textbf{48.89} &\textbf{44.30} \\
        Mistral-7B   & 9.47 &9.36 &8.95 &36.39 &35.36 &34.21 &37.47 &36.38 &35.12 \\
        Qwen2-7B     & 11.60 &12.59 &11.41 &37.99 &39.16 &36.57 &38.96 &40.07 &37.43 \\
        Solar-10.7B  & 14.04 &16.41 &14.44 &40.95 &46.08 &41.45 &41.12 &46.35 &41.66 \\
        GPT3.5-turbo & 11.74 &11.58 &11.09 &39.05 &36.06 &35.51 &39.81 &36.76 &36.13 \\
  \midrule
        Llama3-8B-FT & \textbf{40.56} &\textbf{44.43} &\textbf{41.12} &\textbf{63.11} &\textbf{68.37} &\textbf{63.52} &\textbf{65.33} &\textbf{70.60} &\textbf{65.66} \\
  \bottomrule
\end{tabular}}}
    \caption{Three-shot results of \acp{LLM} on atomic claim decomposition.}
    \label{tab3}
\vspace{-3.0em}
\end{table}

\par For three-shot setting, Llama3-8B achieves the best results on all of the three metrics due to its strong capability in following instructions and learning from context. In addition, with carefully selected examples, Solar-10.7B show its excellent performance second only to Llama3-8B. This result, in contrast to the zero-shot setting, indicating that the model's ability ofinstruction following plays a much greater role than its fundamental language ability when presented with just a limited number of examples. However, all of these results are comparatively not satisfactory, indicating that there is significant room for improvement in the performance of \acp{LLM} on this task. 


\par Fine-tuned model shows best performance on atomic claim decomposition task. We split the \ac{CACDD} dataset into training and testing sets, fine tune the model with 350 pieces of data, and test it with 150 pieces of data. The results indicate that fine-tuning with a small amount of high-quality data can significantly enhance the performance of \acp{LLM} on this task.

\section{Conclusion}
To tackle the limitation that previous research overlooks the importance of claim decomposition, we introduce a new Chinese dataset for the atomic claim decomposition task. This is the first Chinese dataset designed for this task under the real-world \ac{LFQA} scenario. Following Bertrand Russell's philosophy of logical atomism, we provide a definition of atomic claims and a pipeline to annotate atomic claim data. Our experimental results show that existing \acp{LLM} still have significant room for improvement on this task. In the future, we intend to build a larger-scale dataset and propose a new approach to enhance the capacity of \acp{LLM} for the atomic claim decomposition task.

\subsubsection{Limitations}
The \ac{CACDD} dataset only contains 500 question-answer pairs, which is too small to comprehensively evaluate the \acp{LLM}. Furthermore, we assume that all facts in the \ac{LLM} generated responses are of interest to the user, but some people may have different views on this assumption. 

\begin{credits}
\subsubsection{\ackname} This work was funded by the National Natural Science Foundation of China (NSFC) under Grants No. 62372431 and 62472408, the Strategic Priority Research Program of the CAS under Grants No. XDB0680102, XDB0680301, the National Key Research and Development Program of China under Grants No. 2023YFA1011602, the Youth Innovation Promotion Association CAS under Grants No. 2021100, the Lenovo-CAS Joint Lab Youth Scientist Project, and the project under Grants No. JCKY2022130C039.

\end{credits}
 
%

%
%
%
\bibliographystyle{splncs04}
\bibliography{mybibliography}
%






\end{document}

%% file: packages.tex
\usepackage[T1]{fontenc}
\usepackage{graphicx}
\usepackage{acronym}

\usepackage{multirow}
\usepackage{booktabs}
\usepackage[titletoc]{appendix}
\usepackage{longtable}
\usepackage{tabularx}

%% file: definitions.tex
\AtBeginDocument{%
  \providecommand\BibTeX{{%
    \normalfont B\kern-0.5em{\scshape i\kern-0.25em b}\kern-0.8em\TeX}}}

\acrodef{LLM}{large language model}
\acrodef{RAG}{retrieval-augmented generation}
\acrodef{CACDD}{Chinese Atomic Claim Decomposition Dataset}
\acrodef{NLTK}{Natural Language Toolkit}
\acrodef{LFQA}{Long-form Question Answering}
\acrodef{NLP}{Natural Language Processing}

%% file: main.bbl
\begin{thebibliography}{10}
\providecommand{\url}[1]{\texttt{#1}}
\providecommand{\urlprefix}{URL }
\providecommand{\doi}[1]{https://doi.org/#1}

\bibitem{chen2024complex}
Chen, J., Kim, G., Sriram, A., Durrett, G., Choi, E.: Complex claim verification with evidence retrieved in the wild. In: Proceedings of the 2024 Conference of the North American Chapter of the Association for Computational Linguistics: Human Language Technologies (Volume 1: Long Papers). pp. 3569--3587 (2024)

\bibitem{chen2022generating}
Chen, J., Sriram, A., Choi, E., Durrett, G.: Generating literal and implied subquestions to fact-check complex claims. In: Proceedings of the Conference on Empirical Methods in Natural Language Processing (EMNLP) (2022)

\bibitem{chen2024sub}
Chen, S., Zhang, H., Chen, T., Zhou, B., Yu, W., Yu, D., Peng, B., Wang, H., Roth, D., Yu, D.: Sub-sentence encoder: Contrastive learning of propositional semantic representations. In: Proceedings of the 2024 Conference of the North American Chapter of the Association for Computational Linguistics: Human Language Technologies (Volume 1: Long Papers). pp. 1596--1609 (2024)

\bibitem{de2021universal}
De~Marneffe, M.C., Manning, C.D., Nivre, J., Zeman, D.: Universal dependencies. Computational linguistics  \textbf{47}(2),  255--308 (2021)

\bibitem{dhuliawala2024chain}
Dhuliawala, S., Komeili, M., Xu, J., Raileanu, R., Li, X., Celikyilmaz, A., Weston, J.E.: Chain-of-verification reduces hallucination in large language models. In: ICLR 2024 Workshop on Reliable and Responsible Foundation Models (2024)

\bibitem{gao2023rarr}
Gao, L., Dai, Z., Pasupat, P., Chen, A., Chaganty, A.T., Fan, Y., Zhao, V., Lao, N., Lee, H., Juan, D.C., et~al.: Rarr: Researching and revising what language models say, using language models. In: Proceedings of the 61st Annual Meeting of the Association for Computational Linguistics (Volume 1: Long Papers). pp. 16477--16508 (2023)

\bibitem{gao2023enabling}
Gao, T., Yen, H., Yu, J., Chen, D.: Enabling large language models to generate text with citations. In: Proceedings of the 2023 Conference on Empirical Methods in Natural Language Processing. pp. 6465--6488 (2023)

\bibitem{gao2019automated}
Gao, Y., Sun, C., Passonneau, R.J.: Automated pyramid summarization evaluation. In: Proceedings of the 23rd Conference on Computational Natural Language Learning (CoNLL) (2019)

\bibitem{glm2024chatglm}
GLM, T., Zeng, A., Xu, B., Wang, B., Zhang, C., Yin, D., Rojas, D., Feng, G., Zhao, H., Lai, H., Yu, H., Wang, H., Sun, J., Zhang, J., Cheng, J., Gui, J., Tang, J., Zhang, J., Li, J., Zhao, L., Wu, L., Zhong, L., Liu, M., Huang, M., Zhang, P., Zheng, Q., Lu, R., Duan, S., Zhang, S., Cao, S., Yang, S., Tam, W.L., Zhao, W., Liu, X., Xia, X., Zhang, X., Gu, X., Lv, X., Liu, X., Liu, X., Yang, X., Song, X., Zhang, X., An, Y., Xu, Y., Niu, Y., Yang, Y., Li, Y., Bai, Y., Dong, Y., Qi, Z., Wang, Z., Yang, Z., Du, Z., Hou, Z., Wang, Z.: Chatglm: A family of large language models from glm-130b to glm-4 all tools (2024)

\bibitem{hassan2015detecting}
Hassan, N., Li, C., Tremayne, M.: Detecting check-worthy factual claims in presidential debates. In: Proceedings of the 24th acm international on conference on information and knowledge management. pp. 1835--1838 (2015)

\bibitem{hu2024benchmarking}
Hu, N., Chen, J., Wu, Y., Qi, G., Bi, S., Wu, T., Pan, J.Z.: Benchmarking large language models in complex question answering attribution using knowledge graphs. arXiv preprint arXiv:2401.14640  (2024)

\bibitem{jing2023faithscore}
Jing, L., Li, R., Chen, Y., Jia, M., Du, X.: Faithscore: Evaluating hallucinations in large vision-language models. arXiv preprint arXiv:2311.01477  (2023)

\bibitem{kamoi2023wice}
Kamoi, R., Goyal, T., Rodriguez, J.D., Durrett, G.: Wice: Real-world entailment for claims in wikipedia. In: Proceedings of the 2023 Conference on Empirical Methods in Natural Language Processing. pp. 7561--7583 (2023)

\bibitem{lemmon1967comments}
Lemmon, E.J.: Comments on d. davidson’s “the logical form of action sentences”’. The logic of decision and action pp. 96--103 (1967)

\bibitem{li2024minimal}
Li, X., Chen, S., Kapadia, R., Ouyang, J., Zhang, F.: Minimal evidence group identification for claim verification. arXiv preprint arXiv:2404.15588  (2024)

\bibitem{meta2024introducing}
Meta, A.: Introducing meta llama 3: The most capable openly available llm to date, 2024. URL https://ai. meta. com/blog/meta-llama-3/. Accessed on April  \textbf{26} (2024)

\bibitem{min2023factscore}
Min, S., Krishna, K., Lyu, X., Lewis, M., Yih, W.t., Koh, P., Iyyer, M., Zettlemoyer, L., Hajishirzi, H.: Factscore: Fine-grained atomic evaluation of factual precision in long form text generation. In: Proceedings of the 2023 Conference on Empirical Methods in Natural Language Processing. pp. 12076--12100 (2023)

\bibitem{niklaus2019dissim}
Niklaus, C., Cetto, M., Freitas, A., Handschuh, S.: Dissim: A discourse-aware syntactic text simplification framework for english and german. In: Proceedings of the 12th International Conference on Natural Language Generation. pp. 504--507 (2019)

\bibitem{qin2023webcpm}
Qin, Y., Cai, Z., Jin, D., Yan, L., Liang, S., Zhu, K., Lin, Y., Han, X., Ding, N., Wang, H., et~al.: Webcpm: Interactive web search for chinese long-form question answering. In: Proceedings of the 61st Annual Meeting of the Association for Computational Linguistics (Volume 1: Long Papers). pp. 8968--8988 (2023)

\bibitem{russell1918philosophy}
Russell, B.: The philosophy of logical atomism: Lectures 1-2. The Monist  \textbf{28}(4),  495--527 (1918)

\bibitem{shi2024trusting}
Shi, W., Han, X., Lewis, M., Tsvetkov, Y., Zettlemoyer, L., Yih, W.t.: Trusting your evidence: Hallucinate less with context-aware decoding. In: Proceedings of the 2024 Conference of the North American Chapter of the Association for Computational Linguistics: Human Language Technologies (Volume 2: Short Papers). pp. 783--791 (2024)

\bibitem{wang2023self}
Wang, X., Wei, J., Schuurmans, D., Le, Q.V., Chi, E.H., Narang, S., Chowdhery, A., Zhou, D.: Self-consistency improves chain of thought reasoning in language models. In: The Eleventh International Conference on Learning Representations (2023)

\bibitem{wang2023factcheck}
Wang, Y., Reddy, R.G., Mujahid, Z.M., Arora, A., Rubashevskii, A., Geng, J., Afzal, O.M., Pan, L., Borenstein, N., Pillai, A., et~al.: Factcheck-gpt: End-to-end fine-grained document-level fact-checking and correction of llm output. arXiv preprint arXiv:2311.09000  (2023)

\bibitem{wanner2024closer}
Wanner, M., Ebner, S., Jiang, Z., Dredze, M., Van~Durme, B.: A closer look at claim decomposition. arXiv preprint arXiv:2403.11903  (2024)

\bibitem{white2016universal}
White, A.S., Reisinger, D., Sakaguchi, K., Vieira, T., Zhang, S., Rudinger, R., Rawlins, K., Van~Durme, B.: Universal decompositional semantics on universal dependencies. In: Proceedings of the 2016 Conference on Empirical Methods in Natural Language Processing. pp. 1713--1723 (2016)

\bibitem{yue2023automatic}
Yue, X., Wang, B., Chen, Z., Zhang, K., Su, Y., Sun, H.: Automatic evaluation of attribution by large language models. In: Findings of the Association for Computational Linguistics: EMNLP 2023. pp. 4615--4635 (2023)

\end{thebibliography}
